\documentclass[11pt,a4paper, twocolumn]{article}
\usepackage[hyperref]{acl2019}
\usepackage{amsmath}
\usepackage{booktabs}
\usepackage{graphicx}
\usepackage{enumitem}
\usepackage{times}
\usepackage{float}
\usepackage{latexsym}

\usepackage{url}

\aclfinalcopy % Uncomment this line for the final submission

\title{Insertion-Deletion Transformer}

\author{Laura Ruis\thanks{  Work done during internship at Google} \\
  University of Amsterdam \\
  \texttt{laura.ruis@student.uva.nl} \\\And
  Mitchell Stern \\
  University of California, Berkeley \\
  Google Brain \\
  \texttt{mitchell@berkeley.edu} \\\AND
  Julia Proskurnia \\
  Google Inc.\\
  \texttt{juliapro@google.com} \\\And
  William Chan \\
  Google Brain \\
  \texttt{williamchan@google.com} \\}

\date{}

\begin{document}
\maketitle
\begin{abstract}
We propose the Insertion-Deletion Transformer, a novel transformer-based neural architecture and training method for sequence generation. The model consists of two phases that are executed iteratively, 1) an insertion phase and 2) a deletion phase. The insertion phase parameterizes a distribution of insertions on the current output hypothesis, while the deletion phase parameterizes a distribution of deletions over the current output hypothesis. The training method is a principled and simple algorithm, where the deletion model obtains its signal directly on-policy from the insertion model output. We demonstrate the effectiveness of our Insertion-Deletion Transformer on synthetic translation tasks, obtaining significant BLEU score improvement over an insertion-only model.
\end{abstract}

\section{Introduction and Related Work}

Neural sequence models \cite{sutskever-nips-2014,cho-emnlp-2014} typically generate outputs in an autoregressive left-to-right manner. These models have been successfully applied to a range of task, for example machine translation \cite{vaswani-nips-2017}. They often rely on an encoder that processes the source sequence, and a decoder that generates the output sequence conditioned on the output of the encoder. The decoder will typically generate the target sequence one token at a time, in an autoregressive left-to-right fashion.

Recently, research in insertion-based non- or partially- autoregressive models has spiked \cite{stern-icml-2019,welleck-icml-2019,gu-arxiv-2019,chan-arxiv-2019}. These model are more flexible than their autoregressive counterparts. They can generate sequences in any order, and can benefit from parallel token generation. They can learn complex orderings (e.g., tree orderings) and may be more applicable to task like cloze question answering \cite{chan-arxiv-2019} and text simplification, where the order of generation is not naturally left to right, and the source sequence might not be fully observed. 
One recently proposed approach is the Insertion Transformer \cite{stern-icml-2019}, where the target sequence is modelled with insertion-edits. As opposed to traditional sequence-to-sequence models, the Insertion Transformer can generate sequences in any arbitrary order, where left-to-right is a special case. Additionally, during inference, the model is endowed with parallel token generation capabilities. The Insertion Transformer can be trained to follow a soft balanced binary tree order, thus allowing the model to generate $n$ tokens in $O(\log_2 n)$ iterations.

In this work we propose to generalize this insertion-based framework, we present a framework which emits both insertions and deletions. Our Insertion-Deletion Transformer consists of an insertion phase and a deletion phase that are executed iteratively. The insertion phase follows the typical insertion-based framework \cite{stern-icml-2019}. However, in the deletion phase, we teach the model to do deletions with on-policy training. We sample an input sequence on-policy from the insertion model (with on-policy insertion errors), and teach the deletion model its appropriate deletions.

This insertion-deletion framework allows for flexible sequence generation, parallel token generation and text editing. In a conventional insertion-based model, if the model makes a mistake during generation, this cannot be undone. Introducing the deletion phase makes it possible to undo the mistakes made by the insertion model, since it is trained on the on-policy errors of the insertion phase. The deletion model extension also enables the framework to efficiently handle tasks like text simplification and style transfer by starting the decoding process from the original source sequence. 

A concurrent work was recently proposed, called the Levenshtein Transformer (LevT) \cite{gu-arxiv-2019-levt}. The LevT framework also generates sequences with insertion and deletion operations. Our approach has some important distinctions and can be seen as a simplified version, for both the architecture and the training algorithm.
% The LevT framework uses a training algorithm the authors call `dual policy learning'. The insertion framework learns from the deletion framework and vice versa, and they both also learn from an expert policy.
The training algorithm used in the LevT framework uses an expert policy. This expert policy requires dynamic programming to minimize Levenshtein distance between the current input and the target. This approach was also explored by \citet{dong-acl-2019,sasabour-iclr-2019}. Their learning algorithm arguably adds more complexity than needed over the simple on-policy method we propose. The LevT framework consists of three stages, first the number of tokens to be inserted is predicted, then the actual tokens are predicted, and finally the deletion actions are emitted. The extra classifier to predict the number of tokens needed to be inserted adds an additional Transformer pass to each generation step. In practice, it is also unclear whether the LevT exhibits speedups over an insertion-based model following a balanced binary tree order. In contrast, our Insertion-Deletion framework only has one insertion phase and one deletion phase, without the need to predict the number of tokens needed to be inserted. This greatly simplifies the model architecture, training procedure and inference runtime.

An alternative approach for text editing is proposed by \citet{Xia2017DeliberationNS}, which they dub Deliberation Networks. This work also acknowledges the potential benefits from post-editing output sequences and proposes a two-phase decoding framework to facilitate this.

% KERMIT simplifies this approach by only inserting tokens into the current sequence, effectively reaching a theoretical logarithmic decoding time while getting rid of the placeholder classifier. 

In this paper, we present the insertion-deletion framework as a proof of concept by applying it to two synthetic character-based translation tasks and showing it can significantly increase the BLEU score over the insertion-only framework.%, even if it starts decoding from an empty sequence.

% \cite{dong-acl-2019,sasabour-iclr-2019}

\section{Method}
In this section, we describe our Insertion-Deletion model. We extend the Insertion Transformer \cite{stern-icml-2019}, an insertion-only framework to handle both insertions and deletions.

% \subsection{The Framework}
First, we describe the insertion phase. Given an incomplete (or empty) target sequence $\vec{y}_{t}$ and a permutation of indices representing the generation order $\vec{z}$, the Insertion Transformer generates a sequence of insertion operations that produces a complete output sequence $\vec{y}$ of length $n$. It does this by iteratively extending the current sequence $\vec{y}_{t}$. In parallel inference, the model predicts a token to be inserted at each location $[1, t]$. We denote tokens by $c \in C$, where $C$ represents the vocabulary and locations by $l \in \{1, \dots, |\vec{y}_t|\}$. If the insertion model predicts the special symbol denoting an end-of-sequence, the insertions at that location stop. The insertion model will induce a distribution of insertion edits of content $c$ at location $l$ via $p(c, l | \hat{y}_t)$.

The insertion phase is followed by the deletion phase. The deletion model defines a probability distribution over the entire current hypothesis $\vec{y}_t$, where for each token we capture whether we want to delete it. We define $d \in [0, 1]$, where $d = 0$ denotes the probability of not deleting and $d = 1$ of deleting a token. The model induces a deletion distribution $p(d, l | \vec{y}_t)$ representing whether to delete at each location $l \in [0,  |\vec{y}_t|]$.

One full training iteration consisting of an insertion phase followed by a deletion phase can be represented by the following steps:
\begin{enumerate}[noitemsep]
    \item Sample a generation step $i \sim \text{Uniform}([1, n])$
    \item Sample a partial permutation $z_{1:i-1} \sim p(z_{1:i-1})$ for the first $i - 1$ insertions
    \item Pass this sequence through the insertion model to get the probability distribution over $p(c_i^z \mid x_{1:i-1}^{z, i-1})$ (denote $\hat{x}_t$ short for $x_{1:i-1}^{z, i-1}$).
    \item Insert the predicted tokens into the current sequence $\hat{x}_t$ to get sequence $x_{1:i-1+n^i}^{z, i-1+n^i}$ (where $n^i$ denotes the number of insertions, shorten $x_{1:i-1+n^i}^{z, i-1+n^i}$ by $\hat{x}^*_t$) and pass it through the deletion model.
    \item The output of the deletion model represents the probability distribution $p(d_l \mid l, \hat{x}^*_t) \quad \forall \quad l \in \{1, \dots, t\}$
\end{enumerate}
% add explanations on c, l, C, \hat{x}*

\subsection{Learning}
\begin{figure*}[h]
\includegraphics[width=0.9\textwidth]{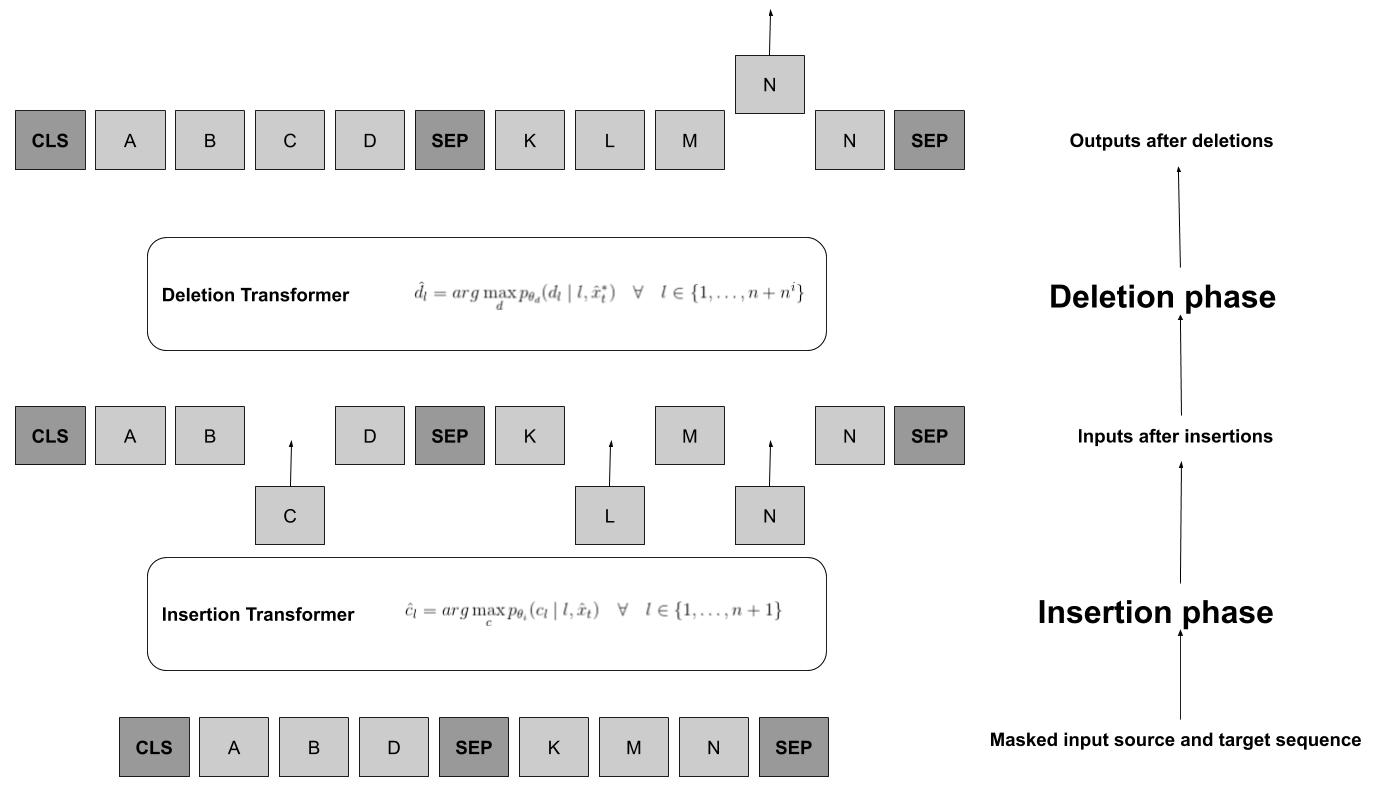}
\caption{Insertion-Deletion Transformer; reads from bottom to top. The bottom row are the source and target sequence, as sampled according to step 1 and 2 in Section 2.1. These are passed through the models to create an output sequence. {\small [CLS] and [SEP] are separator tokens, described in more detail in the BERT paper \cite{bert-paper-2018}. Note that allowing insertions on the input side is not necessary but trains a model that can be conditioned on the input sequence to generate the target as well as vice versa. For details refer to \cite{chan-arxiv-2019}}}
\label{fig:InsDelTransformer}
\end{figure*}

We parametrize both the insertion and deletion probability distributions with two stacked transformer decoders, where $\theta_i$ denotes the parameters of the insertion model and $\theta_d$ of the deletion model. The models are trained at the same time, where the deletion model's signal is dependent on the state of the current insertion model. For sampling from the insertion model we take the argument that maximizes the probability of the current sequence via parallel decoding:  $\hat{c}_l = \arg\max_{c}p(c, \mid l, \hat{x}_t)$. We do not backpropagate through the sampling process, i.e., the gradient during training can not flow from the output of the deletion model through the insertion model. Both models are trained to maximize the log-probability of their respective distributions.
A graphical depiction of the model is shown in Figure~\ref{fig:InsDelTransformer}. 

Since the signal for the deletion model is dependent on the insertion model's state, it is possible that the deletion model does not receive a learning signal during training. This happens when either the insertion model is too good and never inserts a wrong token, or when the insertion model does not insert anything at all. To mitigate this problem we propose an adversarial sampling method. To ensure that the deletion model always has a signal, with some probability $p_{\text{adv}}$ we mask the ground-truth tokens in the target for the insertion model during training. This has the effect that when selecting the token to insert in the input sequence, before passing it to the deletion model, the insertion model selects the incorrect token it is most confident about. Therefore, the deletion model always has a signal and trains for a situation that it will most likely also encounter during inference.

% Footnote referring to KERMIT for detailed equations.

\section{Experiments}
We demonstrate the capabilities of our Insertion-Deletion model through experiments on synthetic translation datasets. We show how the addition of deletion improves BLEU score, and how the insertion and deletion model interact as shown in Table~\ref{tab:inference}. We found that adversarial deletion training did not improve BLEU scores on these synthetic tasks. However, the adversarial training scheme can still be helpful when the deletion model does not receive a signal during training by sampling from the insertion model alone (i.e., when the insertion-model does not make any errors).

\subsection{Learning shifted alphabetic sequences}
The first task we train the insertion-deletion model on is shifting alphabetic sequences. For generation of data we sample a sequence length $\text{min}_n <= n < \text{max}_n$ from a uniform distribution where $\text{min}_n = 3$ and $\text{max}_n = 10$. We then uniformly sample the starting token and finish the alphabetic sequence until it has length $n$. 
For a sampled $n = 5$ and starting letter $\text{c}$, shifting each letter by $\text{max}_n$ to ensure the source and target have no overlapping sequence, here is one example sequence:
\newline
\textbf{Source} \( c\ d\ e\ f\ g \)
\newline
\textbf{Target} \( m\ n\ o\ p\ q \)

\begin{table*}[t]
\scalebox{0.8}{
\begin{tabular}{l|cccccccccccccccccccc}
 \hline \\ Inputs to insertion model        & [CLS] & e & f & g & h & i & j & k & l & m & [SEP] & p & [\_] & [\_] & [\_] & v & w & [SEP]   \\
 Predicted insertions  & & & & & & & & & & &  & o & q & r & u & u &   \\
 Inputs to deletion model & [CLS] & e & f & g & h & i & j & k & l & m & [SEP] & o & p & q & r & u & u & v & w & [SEP] \\
 Predicted deletions & & & & & & & & & & & & & & & & u \\
 Outputs     & [CLS] & e & f & g & h & i & j & k & l & m & [SEP] & o & p & q & r & u & v & w & [SEP]  \\ \hline 
\end{tabular}}
\caption{Example decoding iteration during inference. Here [\_] denotes a space and insertions are inserted to the left of each token in the target sequence (occurring after [SEP]).}
\label{tab:inference}
\end{table*}

We generate 1000 of examples for training, and evaluate on 100 held-out examples. Table \ref{tab:bleu-sequence-shift} reports our BLEU. We train our models for 200k steps, batch size of 32 and perform no model selection. We see our Insertion-Deletion Transformer model outperforms the Insertion Transformer significantly on this task. One randomly chosen example of the interaction between the insertion and the deletion model during a decoding step is shown in Table \ref{tab:inference}.

\begin{table}[h]
\centering
\begin{tabular}{lc}
\toprule
\textbf{Alphabetic Sequence Shifting} & \textbf{BLEU} \\ \midrule
Insertion Model (KERMIT)              & 70.15         \\
Insertion Deletion Model              & 91.49         \\ \bottomrule
\end{tabular}
\caption{BLEU scores for the sequence shifting task.}
\label{tab:bleu-sequence-shift}
\end{table}

\subsection{Learning Caesar's Cipher}
The shifted alphabetic sequence task should be trivial to solve for a powerful sequence to sequence model implemented with Transformers. The next translation task we teach the model is Caesar's cipher. This is an old encryption method, in which each letter in the source sequence is replaced by a letter some fixed number of positions down the alphabet. The sequences do not need to be in alphabetic order, meaning the diversity of input sequences will be much larger than with the previous task. We again sample a $\text{min}_n <= n < \text{max}_n$, where $\text{min}_n = 3$ and $\text{max}_n = 25$ this time. We shift each letter in the source sequence by $\text{max}_n = 25$. If the sampled $n$ is 5, we randomly sample 5 letters from the alphabet and shift each letter in the target to the left by one character we get the following example:
\newline
\textbf{Source} \( h\ k\ b\ e\ t \)
\newline
\textbf{Target} \( g\ j\ a\ d\ s \)

We generate 100k examples to train on, and evaluate on 1000 held-out examples. We train our models for 200k steps, batch size of 32 and perform no model selection. The table below shows that the deletion model again increases the BLEU score over just the insertion model, by around 2 BLEU points.

\begin{table}[H]
\centering
\begin{tabular}{lc}
\toprule
\textbf{Caesar's Cipher} & \textbf{BLEU} \\
\midrule
Insertion Model (KERMIT)              & 35.55         \\
Insertion Deletion Model              & 37.57         \\ \bottomrule
\end{tabular}
\label{tab:bleu-caesar}
\caption{BLEU scores for the Caesar's cipher task.}
\end{table}

%TODO: write out the diversity of potential sequences
% TODO: put generation examples in appendix
\section{Conclusion}

In this work we proposed the Insertion-Deletion transformer, that can be implemented with a simple stack of two Transformer decoders, where the top deletion transformer layer gets its signal from the bottom insertion transformer. We demonstrated the capabilities of the model on two synthetic data sets and showed that the deletion model can significantly increase the BLEU score on simple tasks by iteratively refining the output sequence via sequences of insertion-deletions.
%This framework is generally and efficiently applicable to tasks where the generation order is not autoregressive and the source sequence is partly unobserved. 
The approach can be applied to tasks with variable length input and output sequences, like machine translation, without any adjustments by allowing the model to perform as many insertion and deletion phases as necessary until a maximum amount of iterations is reached or the model predicted an end-of-sequence token for all locations. In future work, we want to verify the capabilities of the model on non-synthetic data for tasks like machine translation, paraphrasing and style transfer, where in the latter two tasks we can efficiently utilize the model's capability of starting the decoding process from the source sentence and iteratively edit the text.

\bibliography{acl2019}

\begin{thebibliography}{12}
\expandafter\ifx\csname natexlab\endcsname\relax\def\natexlab#1{#1}\fi

\bibitem[{Chan et~al.(2019)Chan, Kitaev, Guu, Stern, and
  Uszkoreit}]{chan-arxiv-2019}
William Chan, Nikita Kitaev, Kelvin Guu, Mitchell Stern, and Jakob Uszkoreit.
  2019.
\newblock {KERMIT: Generative Insertion-Based Modeling for Sequences}.
\newblock In \emph{{arXiv}}.

\bibitem[{Cho et~al.(2014)Cho, van Merrienboer, Gulcehre, Bahdanau, Bougares,
  Schwenk, and Bengio}]{cho-emnlp-2014}
Kyunghyun Cho, Bart van Merrienboer, Caglar Gulcehre, Dzmitry Bahdanau, Fethi
  Bougares, Holger Schwenk, and Yoshua Bengio. 2014.
\newblock {Learning Phrase Representations using RNN Encoder-Decoder for
  Statistical Machine Translation}.
\newblock In \emph{{EMNLP}}.

\bibitem[{Devlin et~al.(2018)Devlin, Chang, Lee, and
  Toutanova}]{bert-paper-2018}
Jacob Devlin, Ming{-}Wei Chang, Kenton Lee, and Kristina Toutanova. 2018.
\newblock {{BERT:} Pre-training of Deep Bidirectional Transformers for Language
  Understanding}.
\newblock In \emph{{CoRR}}.

\bibitem[{Dong et~al.(2019)Dong, Li, Rezagholizadeh, and
  Cheung}]{dong-acl-2019}
Yue Dong, Zichao Li, Mehdi Rezagholizadeh, and Jackie Chi~Kit Cheung. 2019.
\newblock {EditNTS: An Neural Programmer-Interpreter Model for Sentence
  Simplification through Explicit Editing}.
\newblock In \emph{{ACL}}.

\bibitem[{Gu et~al.(2019{\natexlab{a}})Gu, Liu, and Cho}]{gu-arxiv-2019}
Jiatao Gu, Qi~Liu, and Kyunghyun Cho. 2019{\natexlab{a}}.
\newblock {Insertion-based Decoding with Automatically Inferred Generation
  Order}.
\newblock In \emph{{arXiv}}.

\bibitem[{Gu et~al.(2019{\natexlab{b}})Gu, Wang, and Zhao}]{gu-arxiv-2019-levt}
Jiatao Gu, Changhan Wang, and Jake Zhao. 2019{\natexlab{b}}.
\newblock {Levenshtein Transformer}.
\newblock In \emph{{arXiv}}.

\bibitem[{Sabour et~al.(2019)Sabour, Chan, and Norouzi}]{sasabour-iclr-2019}
Sara Sabour, William Chan, and Mohammad Norouzi. 2019.
\newblock {Optimal Completion Distillation for Sequence Learning}.
\newblock In \emph{{ICLR}}.

\bibitem[{Stern et~al.(2019)Stern, Chan, Kiros, and
  Uszkoreit}]{stern-icml-2019}
Mitchell Stern, William Chan, Jamie Kiros, and Jakob Uszkoreit. 2019.
\newblock {Insertion Transformer: Flexible Sequence Generation via Insertion
  Operations}.
\newblock In \emph{{ICML}}.

\bibitem[{Sutskever et~al.(2014)Sutskever, Vinyals, and
  Le}]{sutskever-nips-2014}
Ilya Sutskever, Oriol Vinyals, and Quoc Le. 2014.
\newblock {Sequence to Sequence Learning with Neural Networks}.
\newblock In \emph{{NIPS}}.

\bibitem[{Vaswani et~al.(2017)Vaswani, Shazeer, Parmar, Uszkoreit, Jones,
  Gomez, Kaiser, and Polosukhin}]{vaswani-nips-2017}
Ashish Vaswani, Noam Shazeer, Niki Parmar, Jakob Uszkoreit, Llion Jones,
  Aidan~N. Gomez, Lukasz Kaiser, and Illia Polosukhin. 2017.
\newblock {Attention Is All You Need}.
\newblock In \emph{{NIPS}}.

\bibitem[{Welleck et~al.(2019)Welleck, Brantley, Daume, and
  Cho}]{welleck-icml-2019}
Sean Welleck, Kiante Brantley, Hal Daume, and Kyunghyun Cho. 2019.
\newblock {Non-Monotonic Sequential Text Generation}.
\newblock In \emph{{ICML}}.

\bibitem[{Xia et~al.(2017)Xia, Tian, Wu, Lin, Qin, Yu, and
  Liu}]{Xia2017DeliberationNS}
Yingce Xia, Fei Tian, Lijun Wu, Jianxin Lin, Tao Qin, Nenghai Yu, and Tie-Yan
  Liu. 2017.
\newblock Deliberation networks: Sequence generation beyond one-pass decoding.
\newblock In \emph{NIPS}.

\end{thebibliography}
\bibliographystyle{acl_natbib}

\end{document}